\RequirePackage{fix-cm}
\documentclass[twocolumn]{svjour3}          
\smartqed  

\usepackage{hyperref}
\hypersetup{colorlinks,linkcolor=red,anchorcolor=blue,citecolor=blue,CJKbookmarks=True}
\usepackage{epsfig}
\usepackage{times}
\usepackage{url}            
\usepackage{booktabs}       
\usepackage{nicefrac}       
\usepackage{microtype}      
\usepackage{xspace}
\usepackage{graphicx}
\usepackage{wrapfig}
\usepackage{bm}
\usepackage{tabularx}
\usepackage{paralist}
\usepackage{natbib}
\usepackage{bbding}

\usepackage{multirow}%
\usepackage{amsmath,amssymb,amsfonts}%

\usepackage{amsthm}
\usepackage{amsthm}%
\usepackage{mathrsfs}%
\usepackage[title]{appendix}%
\usepackage{xcolor}%
\usepackage{textcomp}%
\usepackage{manyfoot}%
\usepackage{booktabs}%
\usepackage{algorithm}%
\usepackage{algpseudocode}%
\usepackage{listings}%
\usepackage{pifont}

\makeatletter
\DeclareRobustCommand\onedot{\futurelet\@let@token\@onedot}
\def\@onedot{\ifx\@let@token.\else.\null\fi\xspace}

\def\ie{\emph{i.e}\onedot}

\makeatother

\usepackage{colortbl}
\definecolor{tabfirst}{rgb}{1, 0.7, 0.7} 
\definecolor{tabsecond}{rgb}{1, 0.85, 0.7} 
\definecolor{tabthird}{rgb}{1, 1, 0.7} 

\renewcommand{\paragraph}[1]{\vspace{2pt}\noindent {\bfseries #1 \hspace{0.1em}}}
\usepackage{afterpage}

\newcommand{\Tref}[1]{Table~\ref{#1}}

\newcommand{\Aref}[1]{Algorithm~\ref{#1}}
\newcommand{\fref}[1]{Fig.~\ref{#1}}
\newcommand{\eref}[1]{Eq.~\eqref{#1}}
\newcommand{\sref}[1]{Sec.~\ref{#1}}

\usepackage{mathtools}

\newcommand{\set}[1]{\mathbb{#1}} 
\newcommand{\opt}[1]{\mathcal{#1}} 

\newcommand{\sE}{\set{E}}

\newcommand{\sL}{\set{L}}

\newcommand{\oD}{\opt{D}}
\newcommand{\oE}{\opt{E}}
\newcommand{\oF}{\opt{F}}

\newcommand{\oL}{\opt{L}}

\newcommand{\oN}{\opt{N}}

\newcommand{\oU}{\opt{U}}

\begin{document}

\title{
E2VIDiff: Perceptual Events-to-Video Reconstruction\\ using Diffusion Priors
}

\author{Jinxiu Liang$^\dag$, Bohan Yu$^\dag$, Yixin Yang, Yiming Han, and Boxin Shi~\Envelope}

\institute{
$\dag$ Equal contribution.\vspace{0.1cm} \\
\Envelope \, Boxin Shi (Corresponding author)\at
\email{shiboxin@pku.edu.cn} \vspace{0.1cm}\\
\and
Jinxiu Liang \and Bohan Yu \and Yixin Yang \and Boxin Shi\at
National Key Laboratory for Multimedia Information Processing and National Engineering Research Center of Visual Technology, School of Computer Science, Peking University, China. \\
\and
Yiming Han\at
School of Software and Microelectronics, Peking University, China.
}

\date{Received: date / Accepted: date}

\maketitle

\begin{abstract}
Event cameras, mimicking the human retina, capture brightness changes with unparalleled temporal resolution and dynamic range. Integrating events into intensities poses a highly ill-posed challenge, marred by initial condition ambiguities. Traditional regression-based deep learning methods fall short in perceptual quality, offering deterministic and often unrealistic reconstructions. In this paper, we introduce diffusion models to events-to-video reconstruction, achieving colorful, realistic, and perceptually superior video generation from achromatic events. Powered by the image generation ability and knowledge of pretrained diffusion models, the proposed method can achieve a better trade-off between the perception and distortion of the reconstructed frame compared to previous solutions. Extensive experiments on benchmark datasets demonstrate that our approach can produce diverse, realistic frames with faithfulness to the given events.

\end{abstract}

\section{Introduction}
\label{sec:intro}

Known as the `silicon retina', event cameras mimic the human perception system's response to asynchronous, per-pixel logarithmic brightness changes rather than intensity levels periodically. They enjoy unparalleled advantages over conventional frame-based cameras in sensing fast motion of the order of microseconds and a dynamic range larger than 120 dB~\citep{dvs128}, which benefit various machine vision applications~\citep{mitrokhin2018eventbased,vidal2018ultimate}.
Despite their advantages, signals from event cameras have an appearance less intuitive for human perception, which restricts its usage in contexts demanding human-computer collaboration and sophisticated visual analysis—tasks where subjective visual experience is also required and currently beyond the sole capability of machines. This limitation also hinders the integration of advancements in frame-based machine vision, propelled by deep learning and extensive visual data annotations, with event-based machine vision.

Encapsulating the visual signal in a highly compressed form, events hold the potential for `events-to-video' reconstruction with arbitrarily high frame rates and dynamic range \citep{bardow2016simultaneous} to help bridge event cameras to human perception and frame-based machine vision.
However, it is intrinsically an ill-posed problem that can have an infinite number of valid solutions for given events.
Integrating per-pixel brightness changes during a time interval results in only increment images. To obtain the absolute brightness at the end of the interval, an offset image recording the brightness at the beginning of the interval is also needed~\citep{gallego2020eventbased}. 
In addition, the noise model of commercial event cameras deviates from that of conventional cameras. 
Color information is pivotal in human perception, yet the majority of commercially available event cameras primarily capture achromatic (black and white) events.

\begin{figure*}[!t]
	\centering
	\includegraphics[width=0.98\linewidth]{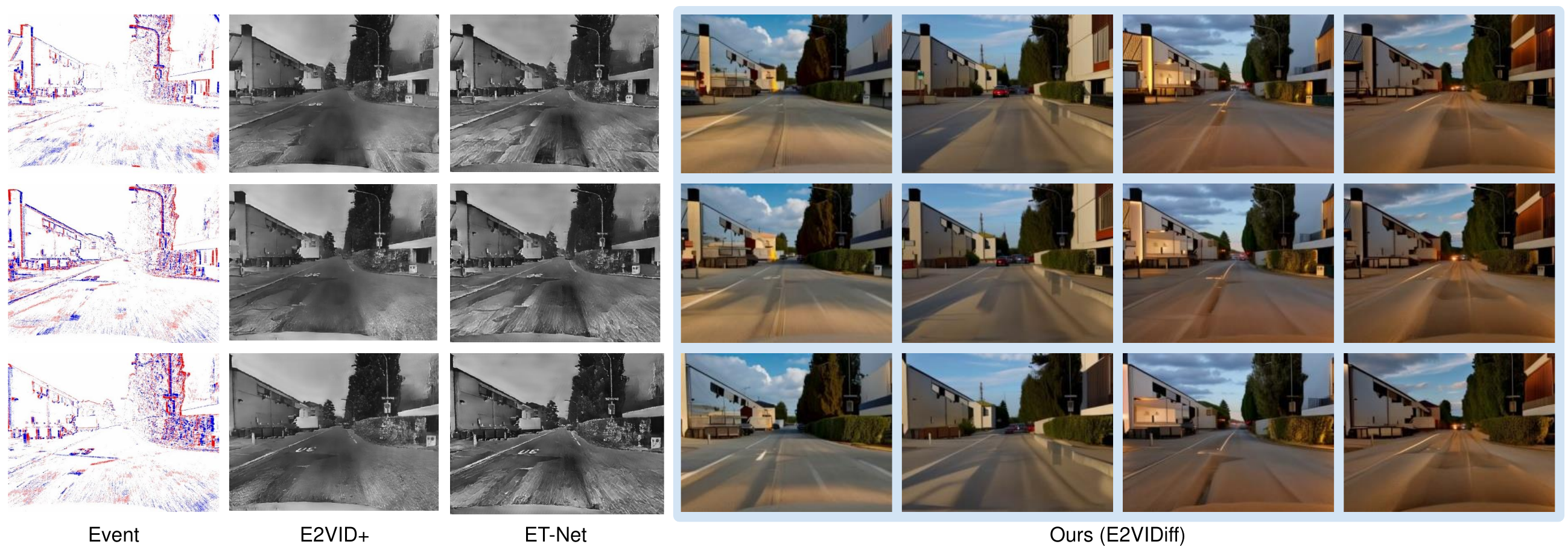}
	\caption{
Visual comparison results of state-of-the-art events-to-video methods E2VID+~\citep{stoffregen2020reducinga}, ETNet~\citep{weng2021eventbased} and the proposed E2VIDiff. 
Given achromatic events, our approach generates high-quality chromatic videos that are perceptually accurate and aligned with the input events, offering photorealism, vivid coloration, and diversity.
Please refer to the supplementary \textit{video} for video reconstruction comparisons.
 }
	\label{fig:fig_diver_dsec_7}
\end{figure*}

Early efforts in converting events to video relied on hand-crafted priors within constrained scene and motion contexts~\citep{kim2014simultaneous,bardow2016simultaneous,munda2018realtime,scheerlinck2019continuoustime}. Recent approaches have shifted towards deep learning paradigms, utilizing ground truth supervision to decrease reliance on restricted assumptions and more effectively capture the intricacies of the real world~\citep{rebecq2019eventstovideo,scheerlinck2020fast,stoffregen2020reducinga,paredes-valles2021back,weng2021eventbased,rebecq2021high,zhu2022eventbased}.
However, these regression-based methods often yield deterministic output, a fundamental mismatch for the inherently ill-posed events-to-video reconstruction challenge. 
They struggle to handle the uncertainty and variability inherent in the unknown initial conditions, noisy event signal, and absent color information.

In situations where several feasible latent frames correspond to the same input events, conventional loss functions often lead reconstructions to converge towards an average of these outcomes, suffering from the \textit{regression to the mean} issue. For example, reconstructing frames of cars driving against diverse street backgrounds from input events can yield multiple valid reconstructions that might reflect cars against streets with varying textures and colors, possibly drawn from Gaussian distributions. Yet, the reconstruction that minimizes the mean squared error typically results in a less detailed, faded, blurred, and low-contrast background - essentially, a conditional mean of latent frames given the events (refer to the roadway and the surrounding vegetation in the results of compared methods in~\fref{fig:fig_diver_dsec_7}).

To address the issue, a promising strategy involves training the model to sample various plausible reconstructions given the events from the posterior distribution, instead of performing point estimations by finding its maximum or calculating its mean.
Recently, diffusion models trained on large-scale dataset become powerful prior models of natural images~\citep{ho2020denoising,rombach2022highresolution} and videos~\citep{guo2023animatediff,luo2023videofusion}, 
which explicitly characterize the conditional data distribution and allow posterior sampling in the image and video manifold.
Despite their unique properties, utilizing pretrained diffusion models to address the aforementioned regression to the mean issue in events-to-video reconstruction is nontrivial, due to the following challenges. First, models trained on large-scale datasets of images and videos have a large \textit{modality gap} with events, which requires massive training with prohibitive expense for effective generalization. However, the current data amount of events is much scarcer than that of images and videos. Second, generative diffusion models risk \textit{compromising faithfulness} to original events, exacerbated by temporal inconsistencies and artifacts inherent in event signals stemming from quantization effects and deviations in the noise model.

In this work, we approach events-to-video reconstruction as a conditional generative modeling problem, introducing \textbf{E}vent-\textbf{To}-\textbf{Vi}deo \textbf{Diff}usion models called \textbf{E2VIDiff} trained to reconstruct latent frames from input events, guided by known ground truth.
Our approaches to overcome challenges in the generative events-to-video process include: 
\textit{i)} employing off-the-shelf video diffusion models pretrained on large-scale dataset, reducing training data requirements while ensuring temporal consistency; 
\textit{ii)} developing a spatiotemporally factorized event encoder and fusion modules for multimodal and temporal integration, which effectively bridge the modality gap; 
\textit{iii)} introducing an event-guided sampling mechanism that directs the denoising process for higher-fidelity reconstructions. 
Experiments demonstrate that this method achieves colorful, photorealistic, and faithful reconstructions from input achromatic events (refer to the results of ours in~\fref{fig:fig_diver_dsec_7}), offering a novel solution for events-to-video reconstruction.
To summarize, our contributions are as follows:
\begin{itemize}
\item pioneering diffusion models for color video reconstruction from achromatic events;
\item efficient modules for bridging the gap between pretrained diffusion models and events; and 
\item an event-guided sampling mechanism for faithful events-to-video reconstructions.
\end{itemize}

\section{Related Work}
\label{sec:related}

\paragraph{Event-to-video reconstruction.}
With the unique advantage of high temporal resolution and high dynamic range, event cameras benefit various vision applications such as tracking~\citep{gehrig2020eklt}, deblurring~\citep{zhang2022eventguided,zhou2023deblurring}, and depth estimation~\citep{rebecq2018emvs,mostafavi2021learning}.
In recent years, there has been a significant focus on image reconstruction from events. 
Early reconstruction approaches relied on hand-crafted priors to estimate the intensities from given events, employing techniques such as
optimization~\citep{bardow2016simultaneous}, regularization~\citep{munda2018realtime}, and temporal filtering~\citep{scheerlinck2019continuoustime,scheerlinck2019asynchronous}.
More recently, deep learning methods have emerged as powerful alternatives for events-to-video reconstruction. 
These methods alleviate the need for specific assumptions about the scene or motion, enabling the handling of more complex scenarios. 
By utilizing
recurrent U-Net architectures~\citep{rebecq2019eventstovideo,rebecq2021high,cadena2021spadee2vid,zou2021learning}, generative adversarial networks (GANs)~\citep{wang2019eventbased}, fully convolutional networks with recurrent layers~\citep{scheerlinck2020fast}, transformer models~\citep{weng2021eventbased}, and spiking neural networks~\citep{zhu2022eventbased}, approaches performing point estimation have demonstrated impressive performance in this regard.
To address the limitation of training data diversity, the use of advanced synthetic datasets has been proposed~\citep{stoffregen2020reducinga}, enhancing the generalization ability of these networks in varied scenes and motions. 
These deep learning approaches mark a significant advancement over traditional methods, offering more robust and adaptable solutions for events-to-video reconstruction. 
However, they tend to produce results with reduced detail, attributed to the regression to the mean issue inherent in point estimators.

\paragraph{Conditional diffusion models.}
Recently, diffusion models have gained significant attention in image modeling due to their robust training objectives and scalability. 
Conditional generation with diffusion models has made notable progress, with approaches that leverage paired data and conditioning information during training. For example, in~\cite{dhariwal2021diffusion}, a classifier is trained in parallel with the denoiser to guide the model in generating images with specific attributes. Similarly,~\cite{choi2021ilvr} conditions the generation on an auxiliary image to guide the denoising process.
There have been works such as~\cite{song2022solving,kawar2021snips,wang2022zeroshot} that utilized pretrained unconditional diffusion models as priors to solve linear inverse imaging problems. These approaches modified the diffusion model sampling algorithm with knowledge of the linear degradation operator to reconstruct images consistent with the learned prior and the given measurements.
Text-to-image synthesis is another active field in diffusion modeling. Works such as~\cite{ramesh2022hierarchicala,nichol2022glide} have pushed the quality of synthesized images to new levels. Latent diffusion models~\citep{rombach2022highresolution} apply diffusion models on lower resolution encoded signals, showing competitive quality with improved speed and efficiency. Imagen~\citep{saharia2022photorealistica} achieves remarkable performance in text-to-image synthesis by using large language models for text processing. Versatile diffusion~\citep{xu2023versatile} further unifies text-to-image, image-to-text, and their variations within a single multiflow diffusion model.
There are fewer works in the domain of video modeling due to the computational cost associated with training on video data and the limited availability of large-scale, publicly available video datasets. Nevertheless, there have been notable contributions, such as~\cite{voleti2022mcvd,khachatryan2023text2videozero,wu2022tuneavideo,blattmann2023align,guo2023animatediff,luo2023videofusion}, which address various aspects of video modeling and synthesis. 
Currently, there are no existing methods that consider the problem of reconstructing color videos from achromatic events.

\section{Method}

\label{sec:method}

In this section, we begin by delineating the problem definition (\sref{sec:problem}) of events-to-video reconstruction. 
Subsequently, an overview of the proposed events-to-video diffusion models E2VIDiff for posterior sampling is presented (\sref{sec:overview}), which precedes a detailed exposition on the reverse sampling strategy to improve faithfulness to given events (\sref{sec:sampling}) and the network architecture tailored to efficiently bridge the modality gap (\sref{sec:network}). 

\begin{figure*}[!t]
	\centering
	\includegraphics[width=0.98\linewidth]{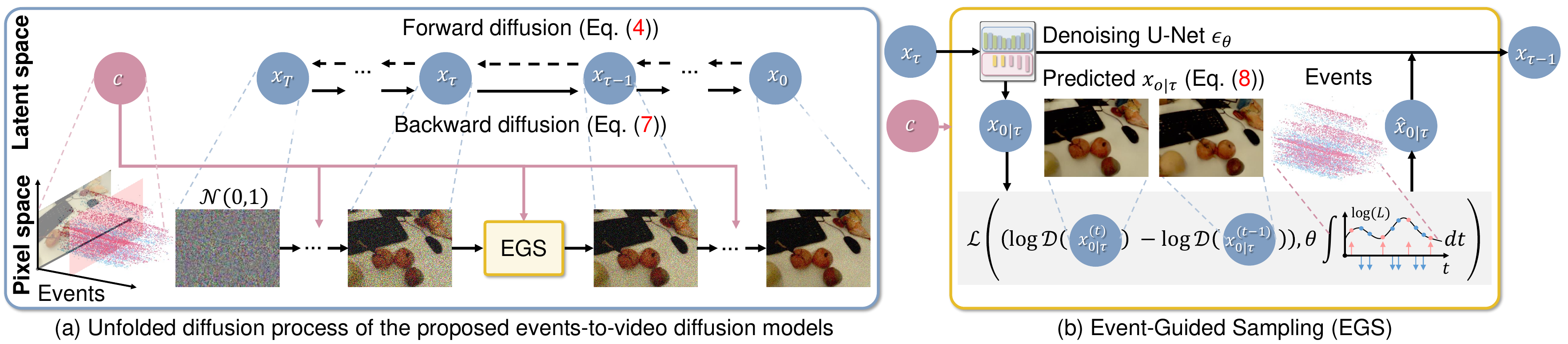}
	\caption{ 
 Schematic of the proposed E2VIDiff. 
(a) The process begins with the latent frame representation $x_0$ derived from frame $L$, which undergoes a forward diffusion via the transition kernel expressed in~\eref{eq:forward}. Conversely, backward diffusion begins from pure Gaussian noise $x_T\sim\oN(0,I)$ and iteratively reconstructs a predicted sample $x_0\sim p(x|c)$ by a backward diffusion expressed in Eqs.~\eqref{eq:ddim} and~\eqref{eq:x0}. 
(b) The proposed event-guided sampling mechanism iteratively enhances the fidelity and perceptual quality of the generated frame by alternating between denoising (to align with natural image characteristics) and ensuring the predicted sample $x_{0|\tau}$ is consistent with the physical formulation of the given events. 
 }
 \label{fig:method}
\end{figure*}

\subsection{Problem definition}
\label{sec:problem}

An event $e_t=(t, j, \sigma)$ at the pixel $j =(j_x,j_y)^\top$ and time $t$ is triggered whenever the logarithmic change of irradiance $L$ exceeds a predefined threshold $\theta$ (\textgreater{} 0),
\begin{equation}
\lVert\log L_t^{(j)} - \log L_{t-\delta t}^{(j)} \rVert \geq \theta,
\label{eq:ev}
\end{equation}
where $L_{t}$ denotes the instantaneous intensity at time $t$, and the polarity $\sigma\in \{-1, +1\}$ indicates
\{negative, positive\} brightness changes. 
Events can be integrated to reconstruct the absolute brightness image of the scene as
\begin{equation}
\log{L_t} = \log{L_{t_{\text{offset}}}} + \theta E_{t_{\text{offset}}\to t},
\label{eq:event_int2}
\end{equation}
where $\log{L_t}$ denotes the logarithmic latent frame, and
\begin{equation}
     E_{t_{\text{offset}}\to t}=\int_{t_\text{offset}}^t{e_\tau d\tau}. 
     \label{eq:evint}
\end{equation}
It can be seen that integration of per-pixel brightness changes during a time interval produces only the brightness \textit{increment image} $E_{t_{\text{offset}}\to t}$, and an \textit{offset image} $L_{t_{\text{offset}}}$ (\ie, the brightness image at the start of the interval) needs to be added to obtain the absolute brightness $L_t$ at the end of the interval. 
Moreover, 
suppose $\widetilde{L}_t$ is a valid solution of frame at timestamp $t$ for certain offset image $\widetilde{L}_{t_{\text{offset}}}$, then $\alpha\widetilde{L}_t$ is also an valid solution corresponds to offset image  $\alpha\widetilde{L}_{t_\text{offset}}$ for any nonzero $\alpha$, including the case where $\alpha$ is spatially varying.
Therefore, solving the events-to-video problem often requires regularization techniques to incorporate prior knowledge of the appearance of a natural image. 
Priors are also necessary to account for quantization effects and noise model deviated from conventional cameras in real-world scenarios, and to prevent overfitting to specific scenes or lighting conditions.

Consider samples ${\sL,\sE}$ drawn from an unknown distribution $p(\sL,\sE)$, where $\sL = \{L^{(t)}\}_{t=0}^{N-1}$ denotes a sequence of $N$ sharp chromatic frames, and $\sE = \{E^{(t)}\}_{t=0}^{N-1}$ denotes a stack of event stream partitioned into $N$ sequential spatiotemporal windows corresponding to the exposure time of the $N$ frames. The goal of events-to-video reconstruction is to reconstruct a sequence of frames $\sL$ from their corresponding event stream $\sE$.
In our work, it is treated as a conditional generative problem, where the conditional distribution $p(\sL|\sE)$ is a one-to-many mapping in which many target images may be consistent with given events.
We are interested in learning a parametric approximation to $p(\sL|\sE)$ through a stochastic translation process that maps given event stream $\sE$ to target frame sequence $\sL$ using diffusion models.

\subsection{Events-to-video diffusion models}

\label{sec:overview}

Our approach leverages diffusion models for posterior sampling in the video manifold, facilitating the reconstruction of diverse, colorful, and perceptually high-quality videos from achromatic input events. 
The overall pipeline of the proposed E2VIDiff is shown in~\fref{fig:method}.

Diffusion models perform data modeling via iterative diffusion and reversely denoising, where the denoising network is trained with denoising score matching~\citep{ho2020denoising}.
To alleviate the computational burden in the pixel space, the proposed E2VIDiff is based on latent diffusion models~\citep{rombach2022highresolution} that operate the diffusion process in the latent space. Specifically,  a compression model is used to transform the frame sequence $\sL \sim p_\text{data}$ into a lower-dimensional latent space. A variational auto-encoder that maximized the lower bound of the evidence~\citep{esser2021taming} leads to the latent presentation $x=\oE(\sL)$, where $\oE$ is the encoder.
The resulting pixel domain representation can be obtained by $\widehat{\sL} = \oD(x)$,
where $\oD$ is the decoder.

During the training phase, the target frame sequence $\sL$ is initially mapped to the latent space via the frozen encoder $\opt{E}$, resulting in $x_{0} = \opt{E}(\sL)$. 
The forward process begins with the representation $x_{{0}}$ that are repeatedly diffused by the transition kernel $q(x_\tau|x_{\tau-1})$ as follows:
\begin{equation}
    q(x_\tau|x_{\tau-1}) = \oN(x_\tau;\sqrt{\alpha_{\tau}}x_{\tau-1}, (1-\alpha_{\tau})I),
    \label{eq:forward}
\end{equation}
where $I$ denotes identity matrix, until it reaches the pure Gaussian noise  $x_T\sim\oN(0,I)$  given a sufficiently large step $T$. 
Here, $0<\alpha_{\tau}<1$ comprises a noise schedule, parameterized by a diffusion step $\tau$, which causes the logarithmic signal-to-noise ratio $\lambda_{\tau} = \log(\alpha_{\tau}/(1-\alpha_{\tau}))$ to decrease monotonically. 
It results in closed-form expressions for a partially noisy latent $x_{\tau}$ at an arbitrary step $\tau$ as:
\begin{equation}
    x_{\tau} = \sqrt{\gamma_{\tau}}x_{0} + \sqrt{1 - \gamma_{\tau}}\epsilon_\tau, \quad \epsilon_\tau \sim \mathcal{N}(0, I),
    \label{eq:forward_x0}
\end{equation}
where $\gamma_\tau=\prod_{j=1}^{\tau}{\alpha_j}$, making the training of a diffusion probabilistic model practical. 
Accordingly, a conditional denoising model $\epsilon_{\theta}(x_{\tau}; c, \tau)$ parameterized by parameters $\theta$ is trained to estimate $\epsilon$ from the diffused $x_{\tau}$ with the conditioning information $c$ extracted from the given event stream $\sE$ as a condition.
The denoising is optimized by minimizing the following $\ell_2$ noise prediction objective:
\begin{equation} 
\mathbb{E}_{x_0 \sim \opt{E}(\sL),\tau\sim p_{\tau},\epsilon_{\tau}\sim \oN(0,I)} [\| \epsilon_{\tau} - \epsilon_\theta(x_{\tau};c,\tau)\|_{2}^{2}],
\label{eq:diffu}
\end{equation}
where $p_{\tau}$ are a uniform distribution over the diffusion time $\tau$.  
The detailed network architecture of $\epsilon_{\theta}(x_{\tau}; c, \tau)$ is introduced in~\sref{sec:network}. 

Upon completion of the training, the reverse process runs backward from pure Gaussian noise $x_T$ to a predicted sample $x_0 \sim p(x|c)$. 
The maximum diffusion time is usually selected so that the input data are completely diffused into Gaussian random noise, which we employ $p_{\tau} \sim \oU\{0, 1000\}$ in the method.
Assuming deterministic DDIM sampling as described by \citet{song2020denoising}, reverse sampling from $p_\theta (x_{\tau-1}|x_\tau)$ is conducted as
\begin{equation}
    x_{\tau-1}=\sqrt{\gamma_{\tau -1}}\widetilde{x}_{0|\tau}+\sqrt{1-\gamma_{\tau -1}} \epsilon_\theta(x_\tau;c, \tau),
    \label{eq:ddim}
\end{equation}
where 
\begin{equation}
    \widetilde{x}_{0|\tau}
    =\left(x_\tau-\sqrt{1-\gamma_\tau}\epsilon_\theta(x_\tau;c, \tau)\right)/\sqrt{\gamma_\tau}.
    \label{eq:x0}
\end{equation}
Denoised estimates of the desired frame sequence on pixel space can be obtained by 
\begin{equation}    \widetilde{\sL}_{\tau}=\oD(\widetilde{x}_{0|\tau}).
    \label{eq:L0}
\end{equation}
To improve faithfulness to given events, we impose the consistency between the predicted consecutive frames $\sL$ and the observed events $\sE$ in the event-guided sampling mechanism detailed in~\sref{sec:sampling}.

\subsection{Event-guided sampling}
\label{sec:sampling}

Generative diffusion models may compromise the faithfulness to given events, often due to temporal discrepancies and artifacts from quantization and deviations in the noise model. Although capable of creating realistic frames, they may produce outputs that deviate significantly from the given events. To mitigate this, we introduce the event-guided sampling (EGS) mechanism, designed to steer the denoising process towards high-fidelity video reconstructions.
Specifically, by using the consistency between the predicted frames $\sL$ and the given events $\sE$ as guidance, EGS iteratively balances the distortion and perceptual quality of the generated frames and progressively projects the samples into the video manifold.
The process of EGS is shown in the right of~\fref{fig:method}.

\begin{algorithm}[t!]
\caption{Event-guided sampling}
\label{alg:egs}
\begin{algorithmic}[1]
\renewcommand{\algorithmicrequire}{\textbf{Required:}}
\Require Encoder $\oE$, decoder $\oD$, U-Net $\epsilon_\theta$, event encoder $\oE_\text{E}$
\renewcommand{\algorithmicrequire}{\textbf{Input:}}
\Require Maximum diffusion time $T$, number of optimization step $S$, number of frame in a sequence $N$, event stack sequence $\{E^{(t)}\}_{t=1}^{N}$, noise schedule $\{\gamma_\tau\}_{\tau=0}^T$
\renewcommand{\algorithmicensure}{\textbf{Output:}}
\Ensure $\{\oD(x_{0}^{(t)})\}_{t=0}^N$
\For{$t = 0$ to $N$}
    \State $x_T^{(t)}\sim\oN(0,I)$
    \State $c^{(t)}=\oE(E^{(t)})$
\EndFor
\For{$\tau = T$ to $1$}
        \For{$t = 0$ to $N$}
            \State $\widehat{x}_\tau^{(t)} \gets x_\tau^{(t)}$
            \State $c=\oE_\text{E}(E^{(t)})$
            \State $\widehat{\epsilon}_\tau^{(t)} \gets \epsilon_\theta(x_\tau^{(t)};c^{(t)}, \tau)$ 
            \For{$s = 0$ to $S-1$}
                \State $\widetilde{x}_{0|\tau}^{(t)}  \gets  \left(x_\tau^{(t)}-\sqrt{1-\gamma_\tau}\widehat{\epsilon}_\tau^{(t)}\right)/\sqrt{\gamma_\tau}$ \Comment{\eref{eq:x0}}
                \State $\widetilde{L}_{\tau}^{(t)}  \gets \oD(\widetilde{x}_{0|\tau}^{(t)})$ \Comment{\eref{eq:L0}}
                \If{$t \ge 1$}
                    \State $d_{\tau}^{(t)}  \gets \log{\widetilde{L}_{\tau}^{(t)}} - \log{\widetilde{L}_{\tau}^{(t-1)}} - \theta E^{(t)}$  
                    \State \Comment{\eref{eq:distance}}
                    \State $\oL \gets  \max \left(0, \Vert d_{\tau}^{(t)}\Vert_1 - 2 \theta\right)  + \eta\Vert x_\tau^{(t)} - \widehat{x}_\tau^{(t)} \Vert_2^2  $ 
                    \State \Comment{Eqs.~\eqref{eq:loss_event},~\eqref{eq:loss_diff} and ~\eqref{eq:loss}}
                    \State ${x}_{\tau}^{(t)} \gets \operatorname{LBFGS}(\oL,x_{\tau}^{(t)})$ 
                \EndIf
            \EndFor
        \State $\widetilde{x}_{0|\tau}^{(t)}  \gets  \left(x_\tau^{(t)}-\sqrt{1-\gamma_\tau}\widehat{\epsilon}_\tau^{(t)}\right)/\sqrt{\gamma_\tau}$ \Comment{\eref{eq:x0}}
        \State $x_{\tau-1}^{(t)} \gets \sqrt{\gamma_{\tau -1}}\widetilde{x}_{0|\tau}^{(t)}+\sqrt{1-\gamma_{\tau -1}} \widehat{\epsilon}_\tau^{(t)}$ \Comment{\eref{eq:ddim}}
    \EndFor
\EndFor
\end{algorithmic}
\end{algorithm}

\begin{figure*}[!t]
    \centering
    \includegraphics[scale=0.2, width=0.98\linewidth]{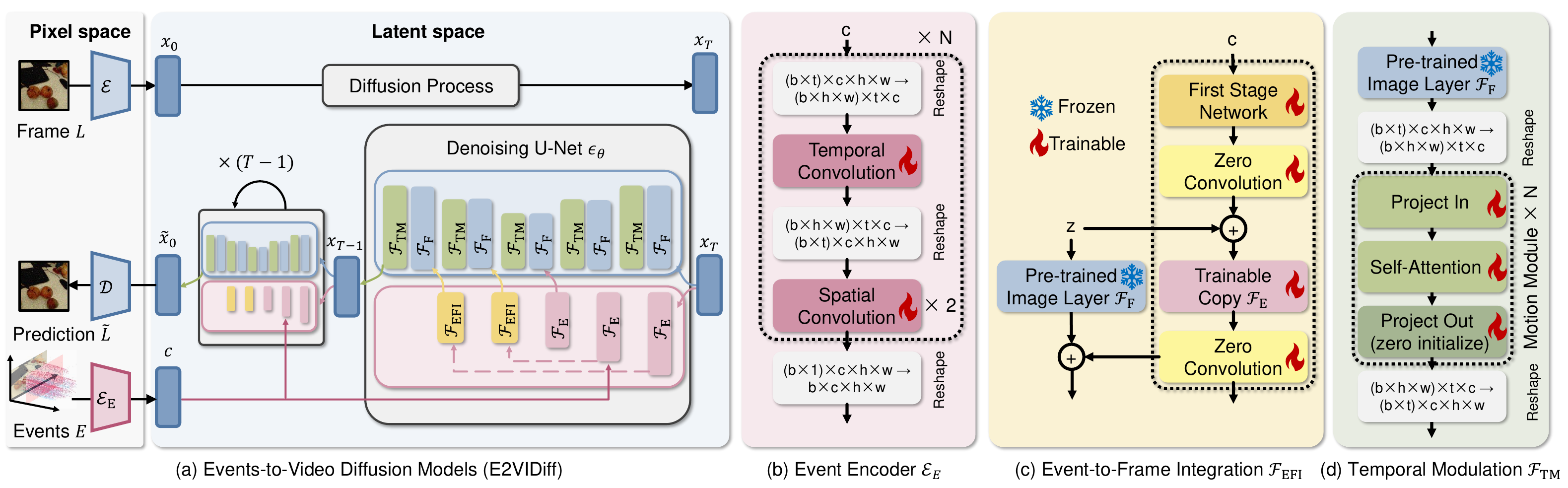}
    \caption{
The network architecture of the denoising U-Net $\epsilon_\theta$ in the proposed E2VIDiff. 
(a) Overview of the denoising process utilizing the denoising U-Net $\epsilon_\theta$, featuring dual branches with weights initialized from pretrained image diffusion models.
(b) The event extractor $\oE_\text{E}$ is designed to extract conditioning information from the event stream.
(c) The event-to-frame integration module $\oF_\text{EFI}$ is designed to merge event features with the output of the frozen image diffusion models.
(d) The temporal modulation module $\oF_\text{TM}$ ensures temporal consistency in the generated frames.
}
\label{fig:net}
\end{figure*} 

Consider the event stack $E^{(t)}$ integrated via~\eref{eq:evint} from the event stream $\{e_k=(t_k, j_k, \sigma_k)\}_{k=0}^K$ with threshold $\theta$ triggered during the exposure time between adjacent frames with indexes $t-1$ and $t$.
Denote the distance between $E^{(t)}$ and the brightness difference between consecutive frames  $\widetilde{L}_{\tau}^{(\cdot)}$ obtained from $x_\tau^{(\cdot)}$ by Eqs.~\eqref{eq:x0} and~\eqref{eq:L0} as:
\begin{equation}
    d_{\tau}^{(t)} = \log{\widetilde{L}_{\tau}^{(t)}} - \log{\widetilde{L}_{\tau}^{(t-1)}} - \theta E^{(t)}.
\label{eq:distance}
\end{equation}
According to~\eref{eq:ev}, a feasible solution satisfies:
\begin{equation}
   -2\theta \leq d_{\tau}^{(t)} \leq 2\theta,
    \label{eq:cons}
\end{equation}
which can be reformulated into the following loss function where the gradient with respect to $x_{\tau}^{(t)}$ can be computed: 
\begin{equation}
\begin{split}
    \oL_\text{event}= \max (0,\Vert d_{\tau}^{(t)} \Vert_1 - 2 \theta).
\end{split}    
\label{eq:loss_event}
\end{equation}
This loss function imposes an $\ell_1$ regularizations on the logarithmic brightness difference between consecutive frames $\sL$ indexed by $t=0,1, ...,N$ to push the solution into the feasible region that consistent to the observation of events $\sE$.

Moreover, to avoid that the solution getting too far away from the feasible space of diffusion model, an $\ell_2$ loss between the guided sampling prediction ${x}_\tau^{(t)}$ and the solution $\widehat{x}_\tau^{(t)}$ sampled from the last DDIM step:
\begin{equation}
    \oL_\text{diff} = {\Vert x_\tau^{(t)} - \widehat{x}_\tau^{(t)} \Vert_2^2}.
\label{eq:loss_diff}
\end{equation}

At each timestamp $\tau$, the overall constraints are formulated as follows:
\begin{equation}
    \oL= \oL_\text{event} + \eta \oL_\text{diff},
    \label{eq:loss}
\end{equation}
where $\eta$ is to balance the weight of different terms, and $x_\tau^{(t)}$ is updated under the guidance of $\oL$ by the Limited-memory Broyden–Fletcher–Goldfarb–Shanno (L-BFGS)~\citep{liu1989limited} optimizer for $S$ steps. 
Under this EGS mechanism, the reverse process~\eref{eq:ddim} is performed iteratively to progressively remove noise and obtain denoised results that are realistic and faithful to the given events.
\Aref{alg:egs} outlines such EGS mechanism.

\subsection{Network architecture}
\label{sec:network}

To mitigate the modality gap of pretrained diffusion models with events, and the limited availability of event data, we develop a spatiotemporally factorized event encoder (EE). Together with an event-to-frame integration (EFI) module, the proposed method effectively bridges the gap between different data modalities. 
Temporal modulation (TM) module is employed to efficiently enhance temporal consistency.
The overall network architecture designs are shown in~\fref{fig:net}.

The denoising U-Net $\epsilon_\theta$ consists of two branches, both with U-Net-like architectures and weights initialized from pretrained image diffusion models as shown in~\fref{fig:net}(a). The frame branch $\oF_\text{F}(x_\tau;\tau)$ accounts for sampling the features of plausible offset image (corresponding to $L_{t_\text{offset}}$ in~\eref{eq:event_int2}) from the random latent $x_T$ sampled from normal space, while the event branch $\oF_\text{E}(x_\tau; c, \tau)$ accounts for generating the features of increment image (corresponding to $E_{t_\text{offset}\to t}$ in~\eref{eq:event_int2})  from the conditioning information $c$ extracted from the input events by the EE module $\oE_\text{E}$.

To align with the encoder $\oE$ used by the pretrained image diffusion models, the EE module $\oE_\text{E}$ convert the event $E^{(t)}$ into a $64\times 64$ feature $c=\oE_\text{E}(E^{(t)})$ for an input of size $512\times 512$ to match the size of the convolutional layer as shown in~\fref{fig:net}(b), where $c$ represents conditioning information. It contains six space-time-factorized event encoder blocks. Each event encoder block consists of a 1D temporal convolution following each 2D spatial convolution layer. 
This enables efficient sharing of information across spatial and temporal dimensions, avoiding the intensive computational demands of 3D convolutional layers. 
Moreover, it distinctly separates the established 2D convolutional layers that can utilize off-the-shelf encoder for similar modalities from the newly introduced 1D convolutional layers. This separation permits the training of the temporal convolutions anew, while preserving the spatial understanding previously acquired in the pretrained weights of the spatial convolutions. 
For effective initialization, the temporal convolution layer starts as an identity function. This strategy ensures a smooth transition from pretrained layers focused solely on spatial aspects to those handling both spatial and temporal elements.

To efficiently integrate the conditioning information from events to frames, the information from the frame branch $\oF_\text{F}$ and the event branch $\oF_\text{E}(x_\tau; c, \tau)$ are fused by the EFI module $\oF_\text{EFI}$ as shown in~\fref{fig:net}(c).
It aligns the feature of events in the low-dimensional latent space defined by pretrained latent diffusion models for natural frames, bridging the gap between different data modalities. 
The TM module $\oF_\text{TM}$ is employed for temporal consistency in frame generation, requiring 5D tensors for input and undergoing a model inflation process for adaptation from pretrained image diffusion models. As shown in~\fref{fig:net}(d), it transforms 2D layers $\oF_\text{F}$ into pseudo-3D, utilizing self-attention blocks for temporal dependency capture, with zero-initialized output layers for smooth integration. This mechanism ensures motion smoothness and content consistency across generated frames.

\begin{figure*}[!t]
	\centering
	\includegraphics[width=0.98\linewidth]{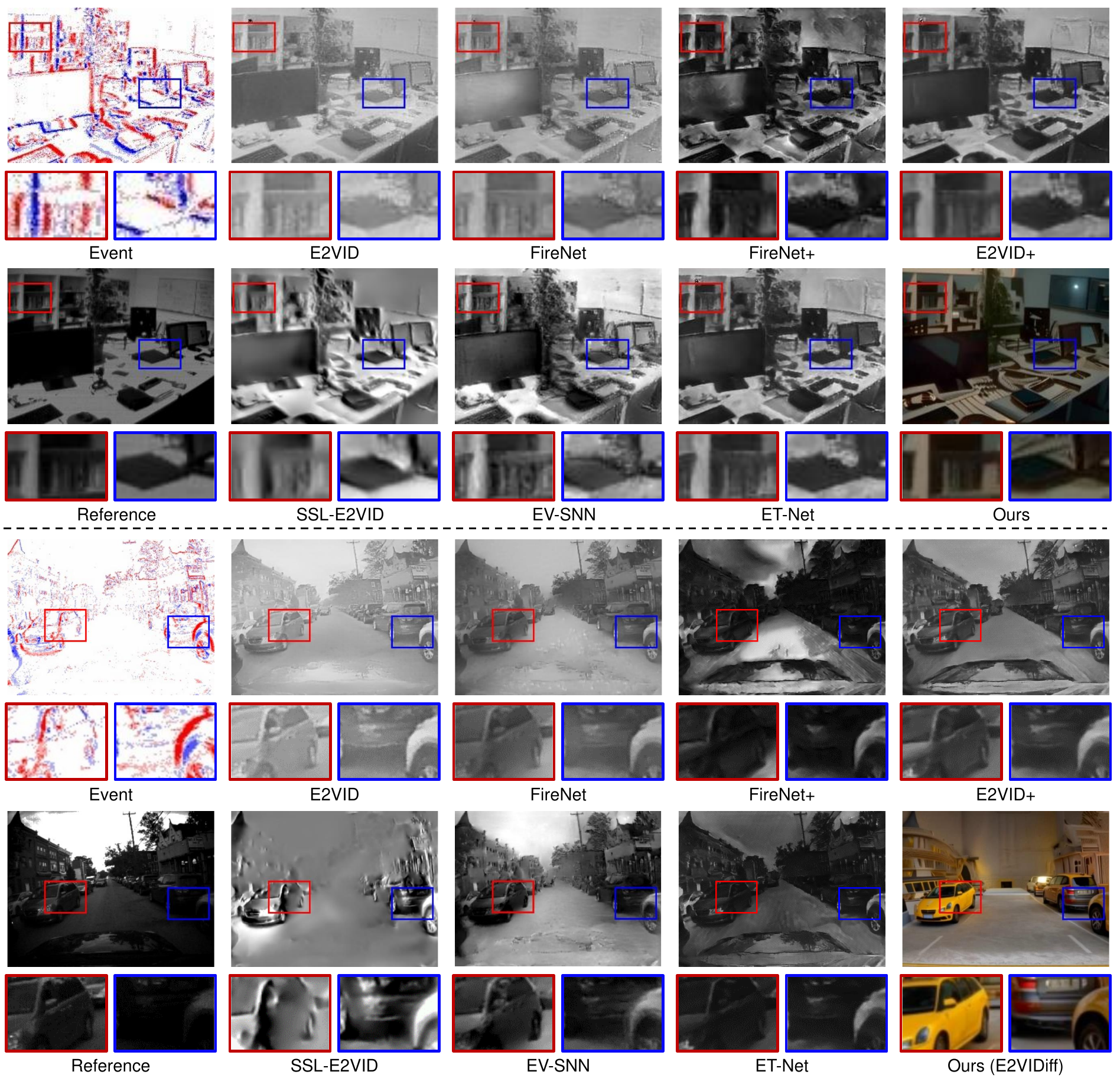}
	\caption{
Qualitative comparisons on real events captured by DAVIS240~\citep{mueggler2017eventcamera} (top) and DAVIS346B~\citep{zhu2018multia} (bottom) event cameras, with corresponding achromatic frames captured concurrently as references. 
The compared state-of-the-art methods include: E2VID~\citep{rebecq2019eventstovideo}, FireNet~\citep{scheerlinck2020fast}, FireNet+~\citep{stoffregen2020reducinga}, E2VID+~\citep{stoffregen2020reducinga}, SSL-E2VID~\citep{paredes-valles2021back}, EV-SNN~\citep{zhu2022eventbased}, and ETNet~\citep{weng2021eventbased}. 
 }
	\label{fig:fig_sota_ijrr2_mvsec1}
\end{figure*}

\begin{figure*}[!t]
	\centering
	\includegraphics[width=0.98\linewidth]{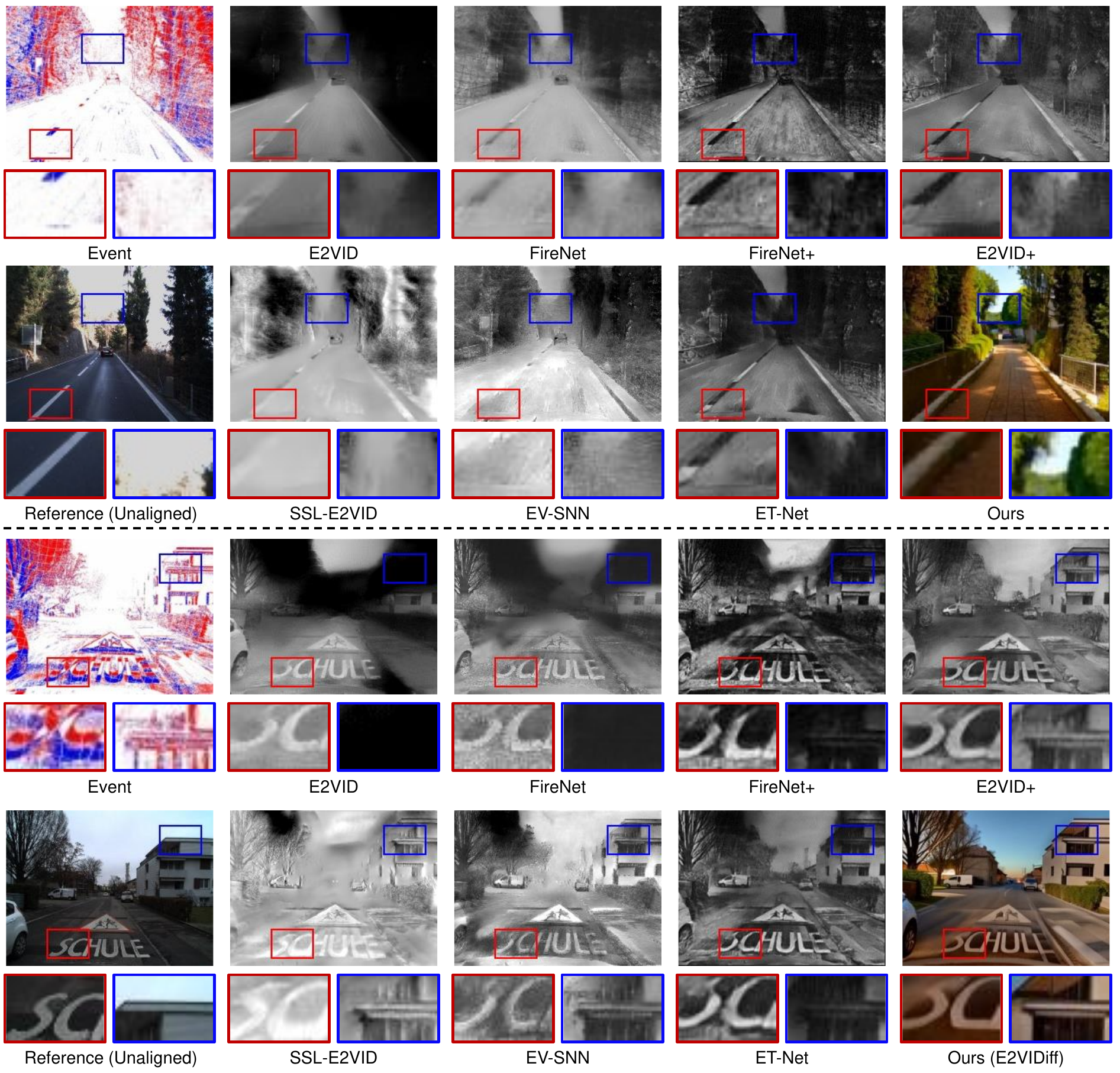}
	\caption{
Qualitative comparisons on real events captured by Prophesee Gen3~\citep{gehrig2021dsec} event cameras, accompanied by chromatic frames captured in stereo alongside the events, serving as unaligned references. 
Please refer to the caption of~\fref{fig:fig_sota_ijrr2_mvsec1} for a complete list of the compared methods. 
}
	\label{fig:fig_sota_dsec2}
\end{figure*}

\begin{figure*}[!t]
	\centering
	\includegraphics[width=0.98\linewidth]{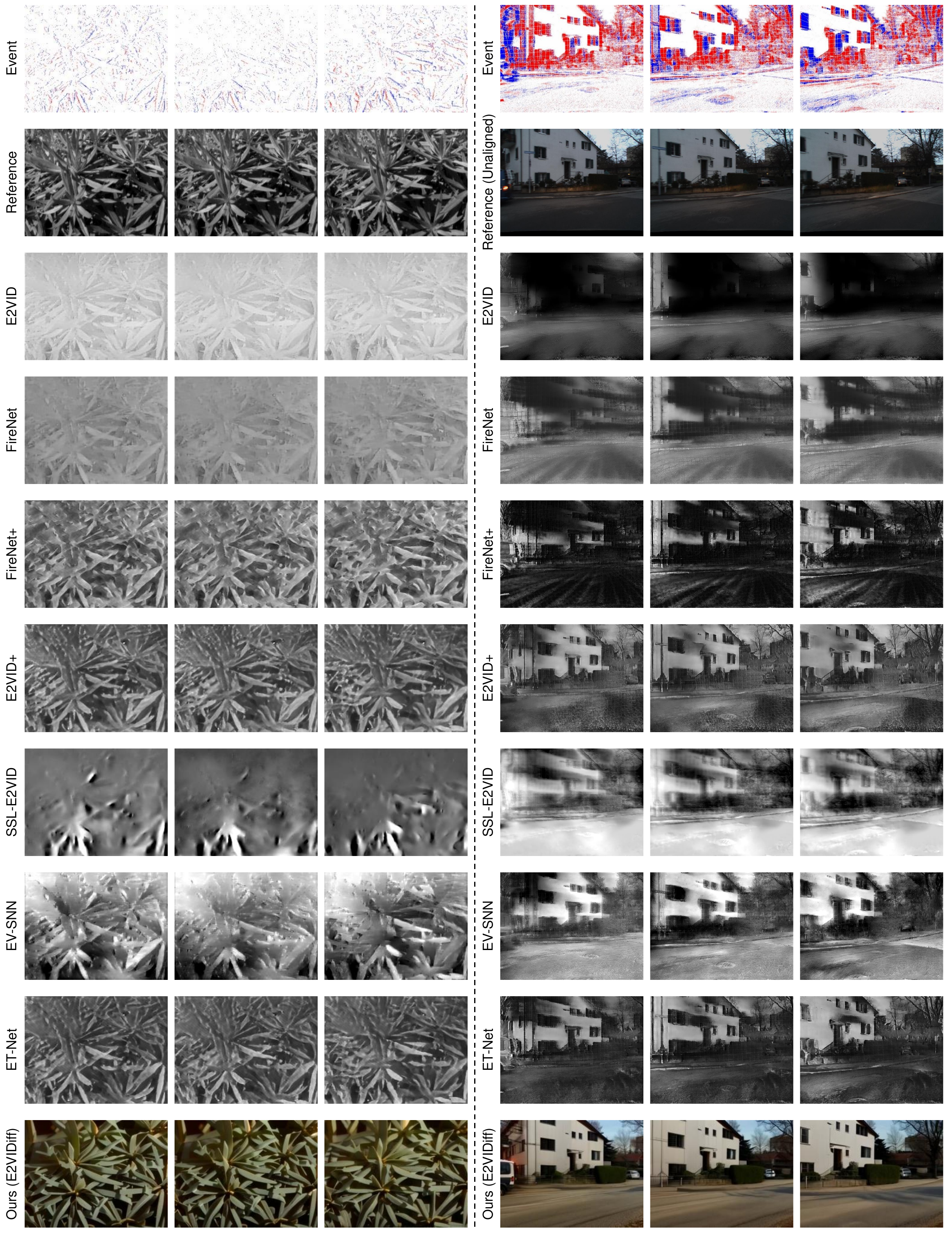}
	\caption{
Qualitative comparisons of temporal consistency in successive frames generated from real events captured by   DAVIS240~\citep{stoffregen2020reducinga} (left) and Prophesee Gen3~\citep{gehrig2021dsec} (right) event cameras, with corresponding achromatic (left) and chromatic (right) frames captured concurrently as references. 
Please refer to the caption of~\fref{fig:fig_sota_ijrr2_mvsec1} for a complete list of the compared methods. 
 }
	\label{fig:fig_temp}
\end{figure*}

\section{Experiments}


\subsection{Experimental setup}

\paragraph{Dataset\footnote{All datasets used for evaluation and training are publicly accessible.}.}
The quantitative evaluation is carried out on three public datasets: HQF~\citep{stoffregen2020reducinga}, IJRR~\citep{mueggler2017eventcamera}, MVSEC~\citep{zhu2018multia}.
The HQF~\citep{stoffregen2020reducinga} dataset comprises 14 sequences captured by the DAVIS240 camera. The IJRR~\citep{mueggler2017eventcamera} dataset includes 25 realistic sequences captured with the DAVIS240 camera. In contrast, the MVSEC~\citep{zhu2018multia} dataset is recorded using a synchronized stereo event camera system. 
All these are real-world event data with the corresponding achromatic frames.
For qualitative evaluation and applications to several high-level downstream tasks, we additionally leverage the DSEC dataset~\citep{gehrig2021dsec} captured using a synchronized stereo event camera system, in which the events are captured by Prophosee EVK3. 
For training, we use a text-video pair dataset WebVid-2.5M~\citep{bain2021frozen}. The video clips in the dataset are sampled at the stride of $5$, then resized and center-cropped to the resolution of $512\times 512$. The event streams corresponding to the video clips are generated from the ideal event generation model~\eqref{eq:ev} first, and augmented with hot pixels and missing event to account for the non-ideal effect of the event triggering.

\paragraph{Metrics.}
Following the evaluation protocol defined in previous works on video diffusion models~\citep{blattmann2023align,luo2023videofusion,yu2023video}, we present the Inception Score (IS)~\citep{salimans2016improved} and Frechet Video Distance (FVD)~\citep{unterthiner2019fvd} scores for various methods, which are pivotal metrics for evaluating the quality and coherence of generated video content in the field of computer vision. 
IS~\citep{salimans2016improved} measures the diversity and clarity of the videos generated. A high IS indicates that the model produces varied and distinct videos, each with clearly defined objects or scenes, as perceived by a trained 3D Convolutional Model (C3D)~\citep{tran2015learninga}. The IS is particularly useful for assessing the generative model's ability to produce visually compelling content that aligns with the distribution of real-world videos.
FVD~\citep{unterthiner2019fvd} evaluates the similarity between the distribution of generated videos and real videos. It can capture both the visual and temporal dynamics of videos, making it a comprehensive measure of video quality and authenticity. A lower FVD score signifies a closer resemblance to real video content, as determined by a trained Inflated 3D ConvNet (I3D) model~\citep{carreira2017quoa}. FVD is crucial for determining the model's effectiveness in capturing the intricacies of video data.

\paragraph{Baselines.}
For comparison, we compare the proposed method with seven state-of-the-art methods: E2VID~\citep{rebecq2019eventstovideo}, FireNet~\citep{scheerlinck2020fast}, E2VID+~\citep{stoffregen2020reducinga}, FireNet+~\citep{stoffregen2020reducinga}, SSL-E2VID~\citep{paredes-valles2021back}, EV-SNN~\citep{zhu2022eventbased}, and ETNet~\citep{weng2021eventbased}. 
All results are generated using the pretrained models provided in the original papers. To ensure strict consistency between the timestamps of the reconstruction and the provided ground truth, we employ the events that occur between two adjacent frames to generate each reconstruction.

\paragraph{Implementation details\footnote{The source code will be released at \url{https://sherrycattt.github.io/E2VIDiff}.}.} The maximum diffusion time $T=1000$. The length of the generated frame sequence $N=16$. The resolution of the video is unrestricted, with a standard resolution of $512 \times 512$. We use a $25$ step DDIM~\citep{song2020denoising} sampler for inference and set the classifier-free guidance scale to $7.5$. A linear beta schedule is used for $\gamma_\tau$ following~\cite{guo2023animatediff}. The number of optimization steps for EGS is $S=5$. To maintain strict timestamp consistency between the reconstruction and ground truth, we utilize the events between two adjacent frames to generate each reconstruction.
The training of the proposed method is explicitly disentangled into three parts, content, structure, and motion. 
The frame branch $\oF_\text{F}$ for the content is initialized from one of the state-of-the-art image diffusion models~\citep{rombach2022highresolution}.  
The most important part is to ensure that each video frame follows the structure or trajectory of the given events. 
In the proposed module, it is achieved by the event branch $\oF_\text{E}$ while freezing the frame branch $\oF_\text{F}$. 
The weights of $\oF_\text{E}$ for per-frame structure-preservation generation are initialized from the HED soft edge variant of~\cite{zhang2023adding}. 
To ensure that the video frames remain temporally coherent, the temporal modulation module $\oF_\text{TM}$ is initialized from the pretrained video diffusion model~\cite{guo2023animatediff}.
The training process takes place on a single NVIDIA GeForce RTX 4090 24GB GPU and spans a duration of 3 days in total. 
The learning rate is $1.0\times 10^{-5}$ and the batch size is $4$.

\begin{table}[!t]
  \centering
  \caption{Quantitative comparison of IS ($\uparrow$) between state-of-the-art methods and the proposed E2VIDiff on different datasets with paired real events and frames. Red, orange, and yellow highlights indicate the 1st, 2nd, and 3rd-best performing method. 
The compared state-of-the-art methods for video reconstruction from events include: E2VID~\citep{rebecq2019eventstovideo}, FireNet~\citep{scheerlinck2020fast}, FireNet+~\citep{stoffregen2020reducinga}, E2VID+~\citep{stoffregen2020reducinga}, SSL-E2VID~\citep{paredes-valles2021back}, EV-SNN~\citep{zhu2022eventbased}, and ETNet~\citep{weng2021eventbased}. 
  }
    \begin{tabular}{lrrr}
  \toprule
    Dataset & \multicolumn{1}{c}{IJRR} & \multicolumn{1}{c}{HQF} & \multicolumn{1}{c}{MVSEC} \\
    \midrule
    FireNet  & \cellcolor{tabsecond} 3.62 &2.79  & \cellcolor{tabthird} 2.92   \\
    FireNet+  & 3.55 & 2.75 & 3.03  \\
    E2VID  & 3.50 & 2.61 &  2.88   \\
    E2VID+  & 3.38& 2.78 &  2.78 \\
    SSL-E2VID  &  \cellcolor{tabthird} 3.54  & \cellcolor{tabfirst} 3.14   &  \cellcolor{tabfirst} 3.13   \\
    EV-SNN  &  3.45  & \cellcolor{tabthird} 2.98   &   2.91   \\
    ETNet  & 3.51 & 2.95 &  2.91 \\
    Ours (E2VIDiff)  & \cellcolor{tabfirst}  3.79 &  \cellcolor{tabsecond} 3.13  &  \cellcolor{tabsecond} 3.06  \\
    \bottomrule
    \end{tabular}%
  \label{tab:IS}%
\end{table}%

\begin{table}[!t]
  \centering
  \caption{Comparison of FVD ($\downarrow$) between state-of-the-art methods and the proposed E2VIDiff on different datasets with paired real events and frames.  Red, orange, and yellow highlights indicate the 1st, 2nd, and 3rd-best performing method. 
Please refer to the caption of~\Tref{tab:IS} for a complete list of the compared methods. 
  }
    \begin{tabular}{lrrr}
  \toprule
    Dataset & \multicolumn{1}{c}{IJRR} & \multicolumn{1}{c}{HQF} & \multicolumn{1}{c}{MVSEC} \\
    \midrule
    FireNet  & 1373.19 & 1614.61 &  2532.41 \\
    FireNet+  & 1438.68 & 1809.96 & \cellcolor{tabsecond} 2159.91 \\
    E2VID  & 1249.41 & 2169.55 & 2622.34  \\
    E2VID+  & \cellcolor{tabfirst} 737.21 & \cellcolor{tabsecond} 974.01 &  2838.63 \\
    SSL-E2VID  &  1799.55  &   3006.62  & \cellcolor{tabthird}  2330.01   \\
    EV-SNN  &  1654.94  &   3177.04  &   2443.08   \\
    ETNet  & \cellcolor{tabsecond} 821.70 & \cellcolor{tabthird} 1193.30 & 2618.33  \\
    Ours (E2VIDiff)  & \cellcolor{tabthird} 885.34 & \cellcolor{tabfirst} 768.03  & \cellcolor{tabfirst} 950.52  \\
    \bottomrule
    \end{tabular}%
  \label{tab:FVD}%
\end{table}%

\subsection{Comparisons with state-of-the-arts}

\paragraph{Qualitative results.}
The visual comparison results are shown in~\fref{fig:fig_sota_ijrr2_mvsec1} and~\fref{fig:fig_sota_dsec2}. 
Thanks to the ability of posterior sampling of diffusion models, the proposed E2VIDiff achieves the best perceptual quality.
This is exemplified by detailed reconstruction of complex scenes, such as the discernible features of bookshelves and the vehicle's head shown in~\fref{fig:fig_sota_ijrr2_mvsec1}, as well as the tree and house shown in~\fref{fig:fig_sota_dsec2}. 
Furthermore, E2VIDiff demonstrates an exceptional ability to accurately infer and assign plausible colors to various scene elements, as observed in the vibrant coloration of the car at the bottom of~\fref{fig:fig_sota_ijrr2_mvsec1}. 
It also excels at capturing scenes characterized by high speed and a wide dynamic range as shown in the driving scenes of~\fref{fig:fig_sota_dsec2}, maintaining clarity and detail where other techniques falter. 
While competing methods struggle to capture the details of the sky, the proposed E2VIDiff retains these elements seamlessly.
Comparatively, while E2VID+~\citep{stoffregen2020reducinga} and ETNet~\citep{weng2021eventbased} emerge as the leading alternatives among the state-of-the-art methods, they exhibit noticeable fog-like artifacts in their results. Furthermore, the inherent limitations of sensor properties and the sparse nature of event steams lead to challenges in generating objects with smooth contours, as evidenced by the jagged contour of the books at the top of~\fref{fig:fig_sota_ijrr2_mvsec1}.

To demonstrate the quality of temporal consistency, we show successive frame results in~\fref{fig:fig_temp}.
Where other methods exhibit various degrees of temporal discontinuities and jitter caused by unstable artifacts, the proposed E2VIDiff maintains a seamless flow, effectively capturing the motion. This is particularly evident in scenes with rapid movement in~\fref{fig:fig_temp}, where our method consistently produces stable and coherent reconstructions. 
The smooth transition between frames, without noticeable artifacts or abrupt changes, highlights the proposed E2VIDiff's proficiency in ensuring temporal consistency, a critical aspect often challenging for video reconstruction tasks.
The fidelity of temporal consistency in our reconstructions can be attributed to the synergistic integration of the EFI module and the EGS mechanism, which work in concert to ensure that each frame not only accurately represents the scene at a given instant, but also coherently aligns with adjacent frames in the temporal domain.

\paragraph{Quantitative results.}
We use the Inception Score (IS)~\citep{salimans2016improved} and Frechet Video Distance (FVD)~\citep{unterthiner2019fvd} scores as the evaluation metrics following~\cite{blattmann2023align,luo2023videofusion,yu2023video}. 
In terms of IS~\citep{salimans2016improved}, SSL-E2VID and the proposed method demonstrates superior performance in terms of IS score, as evidenced in~\Tref{tab:IS}. The good performance can be attributed to our pioneering use of diffusion models for chromatic video reconstruction from achromatic events. By generating samples from the posterior distribution rather than relying on point estimations, our method introduces a level of diversity and clarity in the generated content that significantly enhances the IS score.
For the FVD~\citep{unterthiner2019fvd} scores, our method achieves state-of-the-art performance across all datasets, with the exception of IJRR, as detailed in~\Tref{tab:FVD}. 
This outstanding performance underscores the efficacy of our efficient modules designed to bridge the domain gap between pretrained diffusion models and event data. 
Integration of a spatiotemporally factorized EE module and an EFI module effectively narrows the modality gap, ensuring temporal consistency, and contributing to a lower FVD score. Moreover, our EGS mechanism plays a pivotal role in directing the denoising process, thereby facilitating reconstructions with faithfulness to the given events.

\begin{figure}[!t]
	\centering
\includegraphics[width=0.98\linewidth]{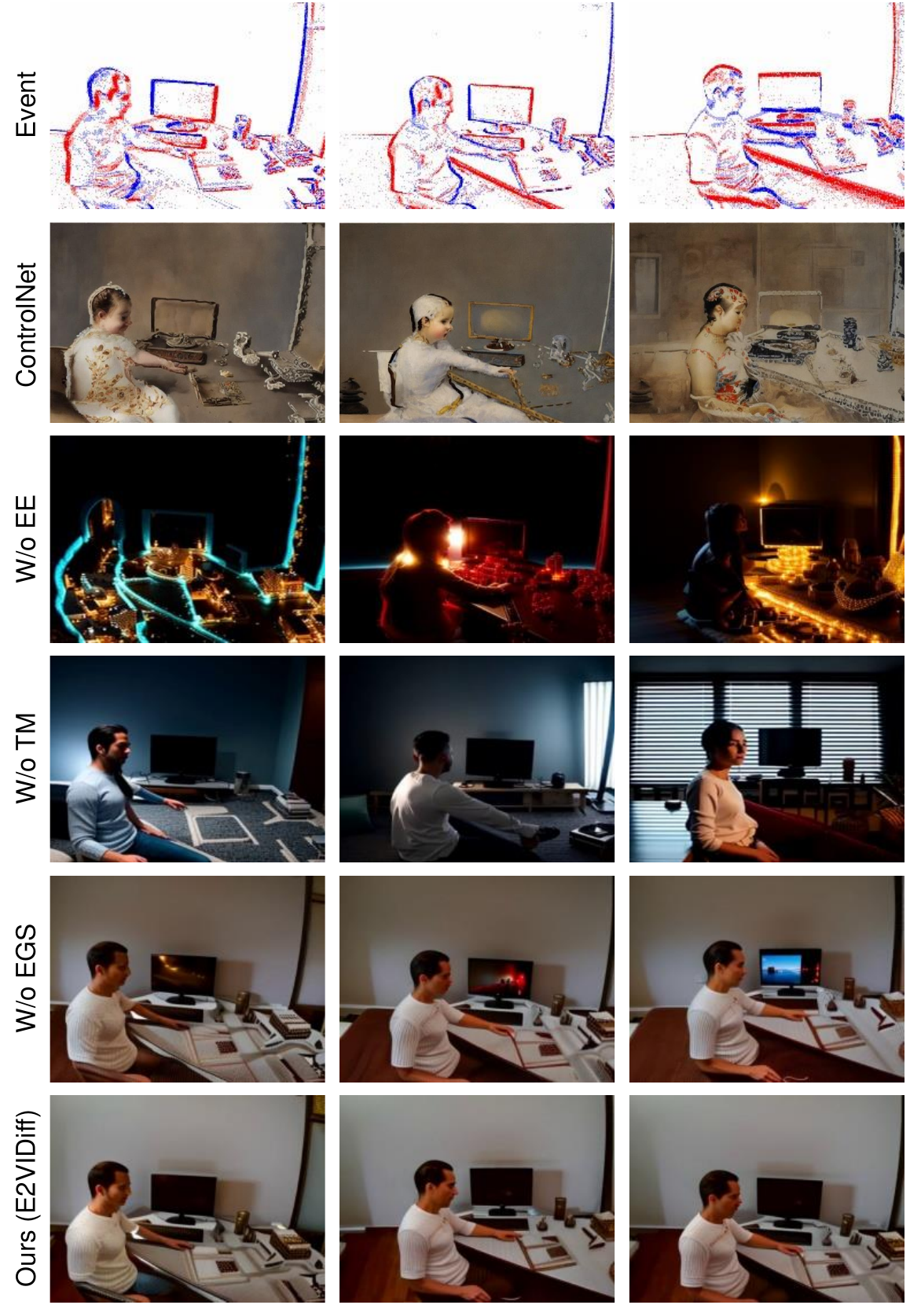}
	\caption{
Ablation study results assessing the individual components of the proposed E2VIDiff. Displayed in sequence from top to bottom are successive frames generated by the following: the baseline ControlNet~\citep{zhang2023adding} with Canny edges derived from integrated events (denoted by `ControlNet'); our method excluding the event encoder (denoted by `W/o EE'); our method without temporal modulation module (denoted by `W/o TM'); our method without the event-guided sampling mechanism (denoted by `W/o EGS'); and the proposed E2VIDiff. 
 }
	\label{fig:ablation}
\end{figure}

\begin{figure*}[!t]
	\centering
	\includegraphics[width=0.9\linewidth]{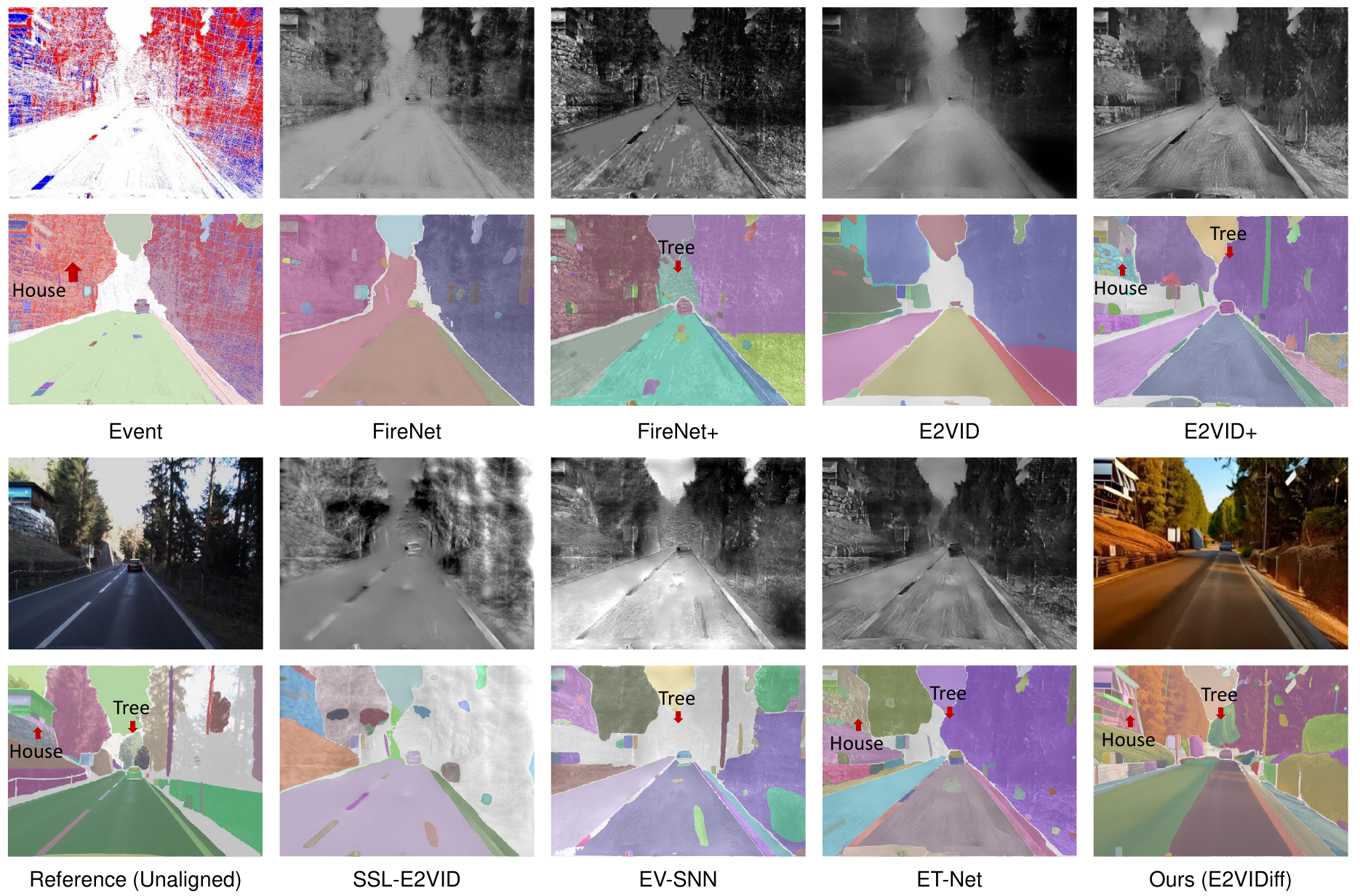}
	\caption{
 Semantic segmentation results employing events-to-video reconstructions as an intermediary, enabling standard computer vision algorithms to be applied on event data directly. All methods utilize the SAM~\citep{kirillov2023segment} for its notable zero-shot performance, suitable for open-domain semantic segmentation on  unaligned reference and frames reconstructed by different methods.
Please refer to the caption of~\fref{fig:fig_sota_ijrr2_mvsec1} for a complete list of the compared methods. 
    }
	\label{fig:downstream_seg}
\end{figure*}

\begin{figure*}[!t]
	\centering
	\includegraphics[width=0.9\linewidth]{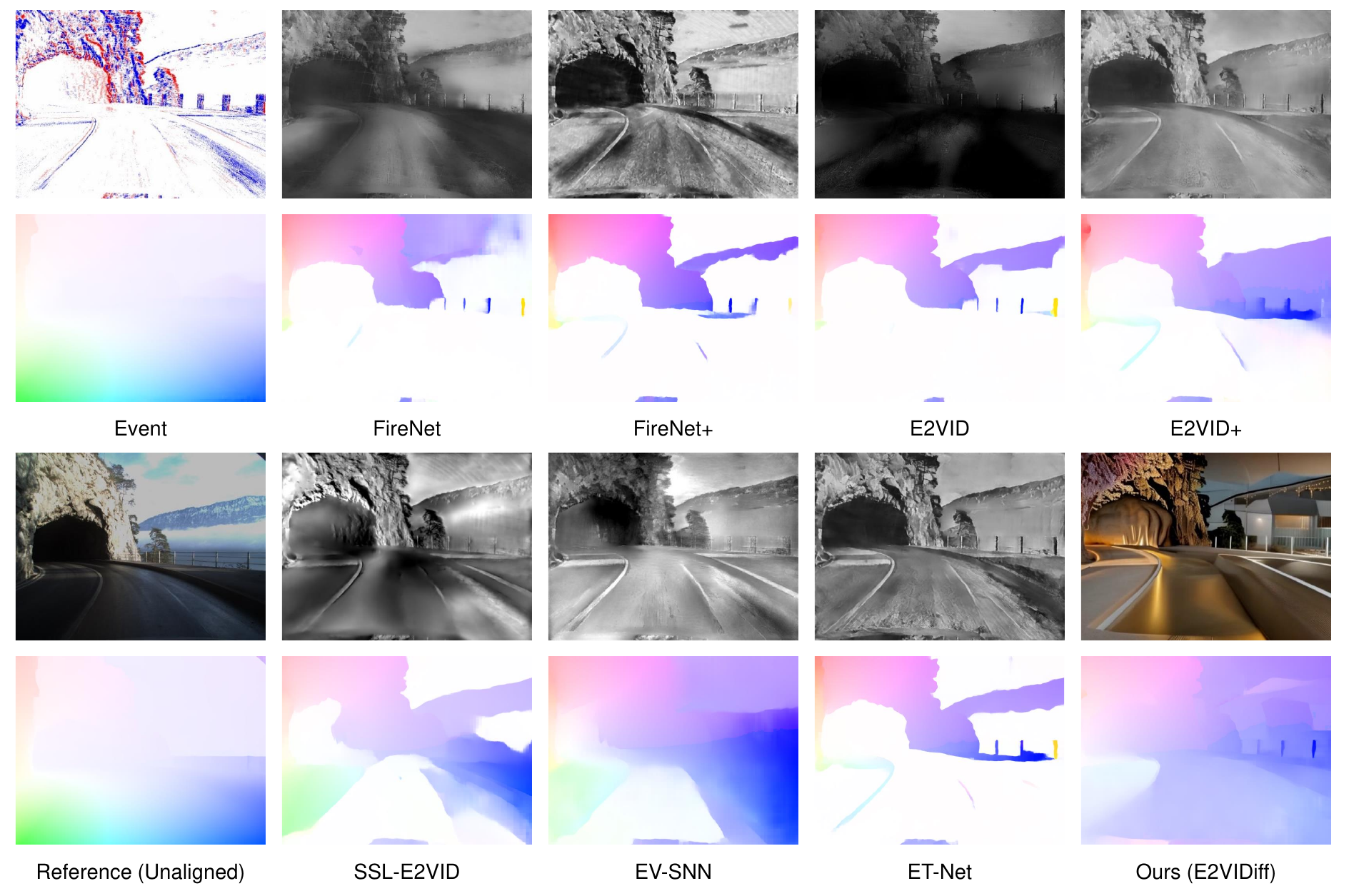}
	\caption{Results of optical flow estimation using events-to-video reconstructions as an intermediary, allowing the direct application of conventional computer vision algorithms to event data. We apply E-RAFT~\citep{gehrig2021eraft} to event data, with comparative results from RAFT~\citep{teed2020raft} on unaligned reference and frames reconstructed by different methods. 
 Please refer to the caption of~\fref{fig:fig_sota_ijrr2_mvsec1} for a complete list of the compared methods. 
 }
	\label{fig:downstream_flow}
\end{figure*}

\subsection{Ablation studies}
In our ablation studies, we evaluate the individual components of our proposed method for chromatic video reconstruction from achromatic events. Qualitative results are illustrated in~\fref{fig:ablation}, showing consecutive frames generated by different configurations of our model. 

The second row presents frames generated by the original ControlNet with Canny edges integrated from events as input, which validate the overall contributions of the proposed components of our method. Integrating Canny edges provides a rudimentary form of event data utilization, which falls short of the sophistication and depth of information leveraged by our proposed method.

The comparison between the frames generated by the model without the EE module and those produced by the proposed E2VIDiff highlights the pivotal role of the EE module in capturing the fine details and dynamics present in the raw events. Without the EE module, the model struggles to accurately interpret and utilize the rich information contained within the events, leading to reconstructions that lack detail and exhibit very poor temporal consistency. The EE module effectively translates the sparse and high-dimensional event data into a more manageable form for the diffusion model, enabling a more precise and coherent video reconstruction.

Omitting the TM module results in frames that fail to maintain consistent motion patterns across time, underscoring the module's crucial role in preserving temporal consistency. The TM module integrates motion information derived from consecutive events, facilitating smooth transitions between frames and enhancing the perception of motion in the reconstructed videos. Its absence leads to a disjointed sequence, where the continuity of movement is compromised, detracting from the overall realism of the video.

The impact of the EGS mechanism is evident when comparing its absence with the proposed E2VIDiff. 
Frames generated without EGS show a noticeable degradation in the fidelity of the reconstructions, with less accurate color representation and diminished clarity. 
The EGS mechanism directs the denoising process in a manner that faithfully adheres to the underlying event data, ensuring that the generated frames not only exhibit high visual quality but also accurately reflect the original scene dynamics.

\subsection{Results of downstream tasks}
We explore using our reconstructions as an intermediary format within the realm of natural images, allowing conventional vision algorithms to be applied directly to events.

\paragraph{Semantic segmentation.}
We perform semantic segmentation tasks to emphasize the precision and reliability of the proposed E2VIDiff. For all the approaches compared, we employ SAM~\citep{kirillov2023segment} for semantic segmentation, which have impressive zero-shot performance that can be applied for open domain. 
For event data, SAM processes on the integrated event stack.
In contrast to reconstructions obtained from~\cite{stoffregen2020reducinga}, our approach yields frames that significantly improve segmentation accuracy. As depicted in~\fref{fig:downstream_seg}, the segmentation of the 'house' region within our reconstructions exhibits a markedly more precise contour, which demonstrates our method's ability to preserve and highlight crucial structural details, thereby enabling segmentation algorithms to perform with greater efficacy.

\paragraph{Optical flow estimation.}
In the realm of optical flow estimation, we employ E-RAFT~\citep{gehrig2021eraft} on event stack, juxtaposed with results generated using RAFT~\citep{teed2020raft} on reconstructed frames. \fref{fig:downstream_flow} illustrates a notable distinction: While other event-to-video reconstruction methods often struggle with flow inference, leading to extensive areas of invalid flow estimation (manifested as white regions), our E2VIDiff achieves comprehensive flow coverage throughout the frame. This capability underscores the method's proficiency in capturing the motion.

\subsection{Sample diversity}
\begin{figure*}[!t]
	\centering
	\includegraphics[width=0.98\linewidth]{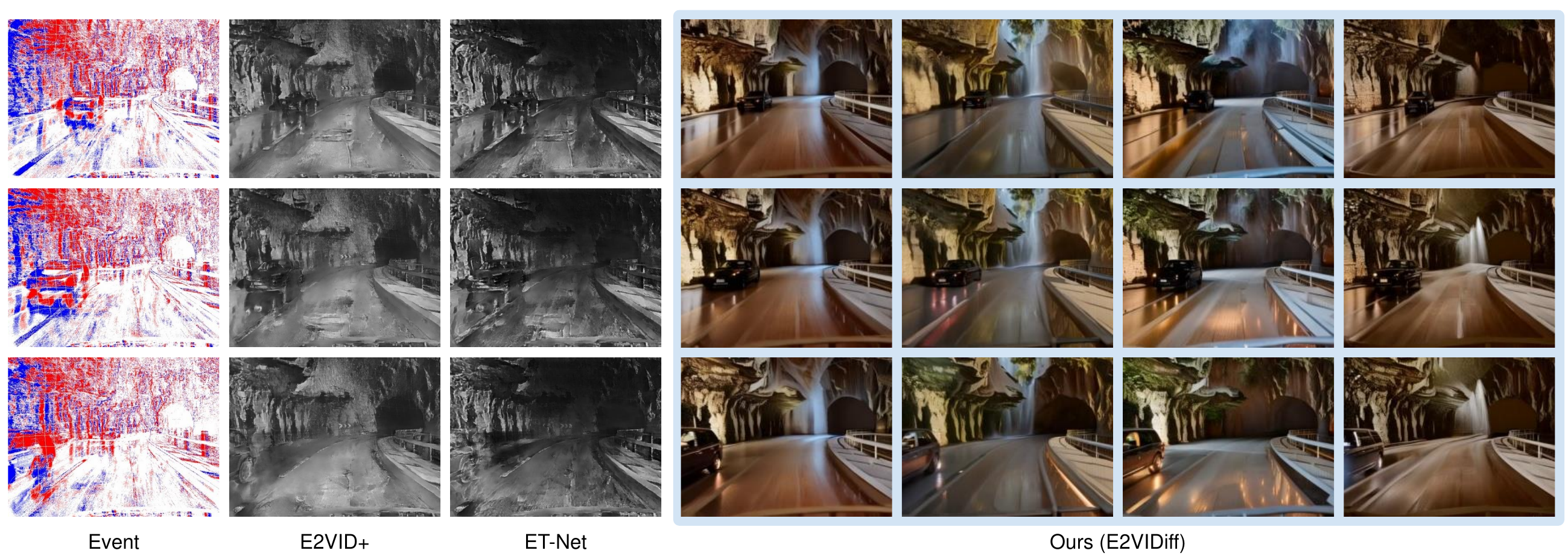}
	\caption{
Qualitative results demonstrating our method's diversity versus E2VID+~\citep{stoffregen2020reducinga} (the second column) and ETNet~\citep{weng2021eventbased} (the third column).
From the input achromatic event stream (first column), our approach yields diverse and plausible chromatic frame reconstructions (final four columns), each depicting distinct aspects of vehicular motion and terrain texture.
}
	\label{fig:fig_diver_dsec8}
\end{figure*}

We present a qualitative analysis that highlights the diversity of the results produced by the proposed E2VIDiff compared to E2VID+~\citep{stoffregen2020reducinga} and ETNet~\citep{weng2021eventbased} in Figs.\ref{fig:fig_diver_dsec_7} and \ref{fig:fig_diver_dsec8}\textcolor{red}{$^\text{4}$}.
\fref{fig:fig_diver_dsec_7} captures an outdoor driving scenario along a street. 
Unlike E2VID+~\citep{stoffregen2020reducinga} and ETNet~\citep{weng2021eventbased}, which tend to produce reconstructions with limited variability, the proposed E2VIDiff excels in generating four distinct outcomes from the same input, where each version reflects unique interpretations of environmental lighting and scene details, thereby illustrating E2VIDiff's adeptness at navigating the uncertainties inherent in real-world scenarios.
\fref{fig:fig_diver_dsec8} shows a car driving along a mountain road. Here, rapid movement and challenging terrain test the temporal consistency and motion interpretation capabilities of the proposed E2VIDiff. 
While E2VID+~\citep{stoffregen2020reducinga} and ETNet~\citep{weng2021eventbased} manage to capture the scene accurately, they offer a singular, constrained view. The proposed E2VIDiff, on the other hand, produces four reconstructions, each articulating different facets of the car's movement and the terrain's roughness. This demonstrates E2VIDiff's exceptional capability to interpret complex motion and terrain dynamics, offering diverse perspectives on fast-motion scenes.

\section{Conclusion}
In this work, we introduce E2VIDiff, leveraging pretrained diffusion models as image and video priors for perceptual events-to-video reconstruction. It tackles the inherent ambiguity in initial conditions and significantly improves the perceptual quality of the reconstructed frames. 
Through the development of an efficient encoder and integration module, we bridge the gap between pretrained diffusion models and events. Furthermore, the introduction of an event-guided sampling mechanism is instrumental in achieving higher fidelity in events-to-video reconstructions. These facilitate the generation of faithful, photorealistic, and chromatic reconstructions from achromatic events, marking an advancement in events-to-video reconstruction tasks.

Although the proposed E2VIDiff has shown remarkable performance in reconstructing video from events in scenarios characterized by high dynamic range and rapid motion (refer to our results shown in Figs.~\ref{fig:fig_sota_dsec2},~\ref{fig:downstream_seg}, and~\ref{fig:downstream_flow}), it falls short in generating high frame rate videos with subtle motion due to the scarcity of such data in the currently publicly accessible large scale datasets for training. 
Additionally, constraints within the motion priors in the pretrained models pose challenges in generating complex motions, such as intricate sports movements or the movement of uncommon objects. These hurdles can be overcome with advances in open source video foundation models that possess more robust motion priors, such as SORA~\citep{brooks2024video}. 
Like other diffusion-based video generation methods, the proposed method also encounters the drawback of extended inference time, a challenge that could be addressed with more efficient sampling strategies. 
Furthermore, while the proposed method aims to harness the capabilities of existing video diffusion models for event-to-video reconstruction, it is necessary to recognize the potential misuse of such technology in generating deceptive content.

\section*{Acknowledgement}
This work was supported by the National Science and Technology Major Project (Grant No. 2021ZD0109803), the
National Natural Science Foundation of China (Grant No. 62302019, 62136001, and 62088102), and the China Postdoctoral Science Foundation (Grant No. 2022M720236).


{\small
\bibliographystyle{spbasic}
\bibliography{ref}
}

\end{document}